# Traditional Machine Learning and Deep Learning Models for Argumentation Mining in Russian Texts

**Fishcheva I. N.**
Vyatka State University,
Kirov, Russia
`fishchevain@gmail.com`

**Goloviznina V. S.**
Vyatka State University,
Kirov, Russia
`golovizninavs@gmail.com`

**Kotelnikov E. V.**
Vyatka State University,
Kirov, Russia;
ITMO University,
Saint Petersburg, Russia
`kotelnikov.ev@gmail.com`

**Abstract**

Argumentation mining is a field of computational linguistics that is devoted to extracting from texts and classifying arguments and relations between them, as well as constructing an argumentative structure. A significant obstacle to research in this area for the Russian language is the lack of annotated Russian-language text corpora. This article explores the possibility of improving the quality of argumentation mining using the extension of the Russian-language version of the Argumentative Microtext Corpus (ArgMicro) based on the machine translation of the Persuasive Essays Corpus (PersEssays). To make it possible to use these two corpora combined, we propose a Joint Argument Annotation Scheme based on the schemes used in ArgMicro and PersEssays. We solve the problem of classifying argumentative discourse units (ADUs) into two classes – "pro" ("for") and "opp" ("against") using traditional machine learning techniques (SVM, Bagging and XGBoost) and a deep neural network (BERT model). An ensemble of XGBoost and BERT models was proposed, which showed the highest performance of ADUs classification for both corpora.

**Keywords:** argumentation mining; machine translation; deep learning; BERT
**DOI:** 10.28995/2075-7182-2021-20-XX-XX

# Модели традиционного машинного обучения и глубокого обучения для анализа аргументации русскоязычных текстов

**Фищева И. Н.**
Вятский государственный университет,
Киров, Россия
`fishchevain@gmail.com`

**Головизнина В. С.**
Вятский государственный университет,
Киров, Россия
`golovizninavs@gmail.com`

**Котельников Е. В.**
Вятский государственный университет,
Киров, Россия;
Университет ИТМО,
Санкт-Петербург, Россия
`kotelnikov.ev@gmail.com`

**Аннотация**

Анализ аргументации – это область компьютерной лингвистики, которая посвящена извлечению из текстов и классификации аргументов и связей между ними, а также построению аргументационной структуры. Существенным препятствием исследованиям в этой области для русского языка является недостаток аннотированных русскоязычных текстовых корпусов. В настоящей статье исследуется возможность повышения качества анализа аргументации при помощи расширения русскоязычной версии Argumentative Microtext Corpus (ArgMicro) на основе машинного перевода Persuasive Essays Corpus (PersEssays). Для возможности совместного применения двух корпусов мы предлагаем объединенную схему разметки на основе схем, используемых в ArgMicro и PersEssays. Мы решаем задачу классификации аргументативных дискурсивных единиц (ADUs) на два класса – "за" и "против" с использованием традиционных методов машинного обучения (SVM, Bagging и XGBoost) и глубокой нейросетевой модели BERT. Был предложен ансамбль моделей XGBoost и BERT, который и показал наивысшее качество классификации ADUs для обоих корпусов.

**Ключевые слова:** анализ аргументации; машинный перевод; глубокое обучение; BERT

# 1 Introduction

Argumentation (or argument) mining is a field of computational linguistics that is devoted to extracting from texts and classifying arguments and relations between them, as well as constructing an argumentation structure [16], [19]. This area is seeing an influx of research activity – for example, since 2014, seven workshops on the analysis of arguments have already been held[1]. Besides being of academic interest, argumentation mining is in the focus of attention due to a wide range of applications, in particular, when studying user opinions based on social media analysis [1], [17], analyzing legal texts [18], scientific texts [9], political debates [25], news articles [3] and student essays [29].

The main text element used in the argumentation mining is an argumentative discourse unit (ADU) – a piece of text that has a single argumentation value [30, p. 63]. As a rule, ADUs are most often individual sentences, but in some cases ADU is a part of a sentence or several sentences.

In ADU-based argumentation mining, the tasks are as follows [30, p. 6]:
1) identifying text fragments containing argumentation;
2) segmenting the text into ADUs;
3) identifying the central (or major) claim (usually among ADUs; but there can also be implicit central claims);
4) classification of ADUs – the main classes are supporting and rebutting ADUs;
5) establishing relations between ADUs;
6) building an argumentation structure;
7) assessing the argumentation quality.

There is also a stance detection task, related to the argumentation mining. This task is to determine the point of view of the text's author and is often solved independently, without identifying arguments [13].

To successfully solve the above mentioned tasks, annotated text corpora are required. Currently, there is a fairly large number of corpora with a variety of argumentative annotation – Lawrence and Reed [16] estimate the known corpora at 2.2 million words. The largest open database of text corpora with argumentative annotation is AifDB[2] [15], which contains more than 14,000 texts. However, most of these corpora are in English.

Fishcheva and Kotelnikov [7] showed that the machine translation of the English-language Argumentative Microtext Corpus (ArgMicro) [24], [27] into Russian allows obtaining the performance of ADUs classification that is not inferior to human translation. In this paper, following [7], we investigate the possibility of improving the performance of ADUs classification based on the extension of the Russian version of the ArgMicro corpus through machine translation of the Persuasive Essays Corpus (PersEssays) [29]. To classify ADUs, we use traditional machine learning techniques (Support Vector Machines – SVM, Bagging and Gradient Boosting – XGBoost implementation), the deep neural network (BERT model [4]), and the XGBoost and BERT ensemble.

When considering argument annotated corpora combined, one of the important problems is the difference in annotation schemes [16]. We propose the Joint Argument Annotation Scheme based on those used in ArgMicro and PersEssays.

The contribution of this paper is as follows:
- the Joint Argument Annotation Scheme that takes into account the peculiarities of ArgMicro and PersEssays annotation schemes is proposed;
- a new Russian-language corpus with argumentative annotation is created. This corpus is formed by expanding the existing Russian-language version of the ArgMicro corpus with the machine translation version of PersEssays corpus. The new corpus is made publicly available;

---

[1] https://argmining2020.i3s.unice.fr.
[2] http://corpora.aifdb.org.

- for the new corpus the performance scores of the ADUs classification into two classes – "pro" and "opp" were obtained based on the traditional machine learning techniques (SVM, Bagging and XGBoost), the neural network model (BERT), as well as the XGBoost and BERT ensemble;
- the effect of expanding the training dataset and the influence of various groups of features on the classification performance was investigated.

The paper is structured as follows. The second section provides an overview of previous work, including existing argument annotation schemes, papers on cross-lingual argumentation mining and argumentation mining for the Russian language. The third section describes the proposed Joint Argument Annotation Scheme. The fourth section is devoted to text corpora and machine learning models for argumentation mining used in this work. In the fifth section the experimental results are presented and discussed. The sixth section provides conclusions and suggests directions for further research.

## 2 Previous work

In this section, firstly, the existing argument annotation schemes are considered, then papers in the field of cross-lingual argumentation mining are given, in conclusion, papers on the Russian-language argumentation mining are indicated.

### 2.1 Argument annotation schemes

The well-known argument annotation schemes are described in [30], as well as in [20]. Almost every corpus uses its own version of the annotation scheme, since different goals were laid down when creating the corpus.

**The microtext scheme** is based on Freeman's theory [8] and is described in detail by Peldszus and Stede [23]. This scheme was used in the annotation of the ArgMicro corpus [24]. In the microtext scheme, an argument structure is seen as a collection of multiple interconnected arguments. A claim (or conclusion) can be supported by premises or attacked by counterarguments. The main types of relations within this scheme are shown in Figure 1.

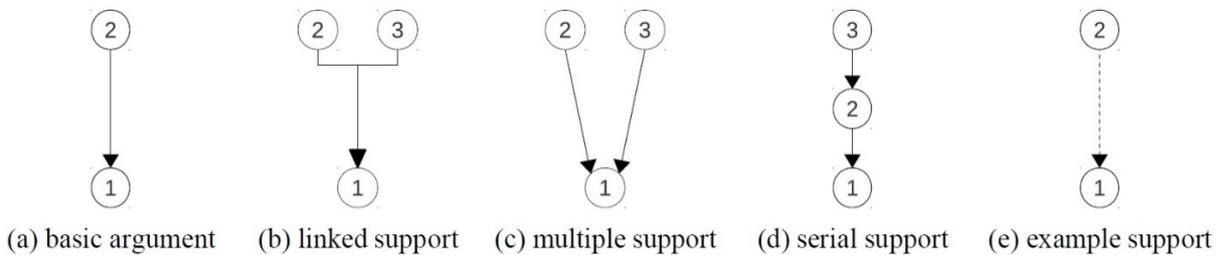

Figure 1: Main types of relations in the microtext scheme [23]

Figure 1 shows the following relations:
- basic argument – one claim is supported by one premise;
- linked support – premises are to be connected before conclusion;
- multiple support – two arguments converging to the same argument;
- serial support – argument can be the premise and the conclusion simultaneously: (2) is a premise for (1), and (2) is a conclusion for (3);
- example support – supporting argument, which is the example.

There are rebutting arguments in this annotation that attack the conclusion or premise. The relation between any arguments can also be questioned (undercut). In this case, rebutting arguments can attack other rebutting arguments.

**The persuasive essay scheme** was used to mark up the PersEssays corpus and is described in detail by Stab and Gurevych [29]. This annotation scheme includes argument components and argumentative relations between the components. One of the argument components called *Major Claim* – it expresses the stance of the essay's author on the topic under discussion. *Claim* arguments support or attack the *Major Claim*. An attribute of the *Claim* ("for" or "against") indicates the polarity related to the *Major Claim*. *Premise* arguments support or attack the *Claims* or another *Premise*. The argumentative relations are defined by the discourse structure.

**The science scheme** is annotation scheme on a fine-grained level in scientific journal articles from the educational domain [11]. It uses a graph of arguments, which links support, attack, detail, and the undirected sequence relations.

**The Modified Toulmin Scheme.** Habernal and Gurevych [10] used the Toulmin model as a basis [32]. They analyzed user-generated web discourse. Within this scheme, there is no need to explicitly annotate any relations between the nodes.

**The Cornell eRulemaking Scheme.** Niculae et al. [21] considered a corpus of user comments on government rule making. It turned out that a lot of comments in the corpus could not be tagged using the microtext or the persuasive essay schemes. Therefore, specific argumentative role labels and new relation types were introduced.

Within our study, we use two existing corpora – ArgMicro and PersEssays. To use them together, we propose the Joint Argument Annotation Scheme based on microtext and persuasive essay schemes.

### 2.2 Cross-lingual argumentation mining

Currently, due to quite a large array of English-language text corpora, annotated by argumentation, and at the same time, the lack of such corpora for other languages, a range of works has emerged where the argumentation annotation of a corpus in one language is used to annotate a corpus in another language.

The current situation, when there is a sufficiently large array of English-language text corpora, annotated by argumentation, and at the same time, there is a lack of such corpora for other languages, has led to the emergence of a range of works on transfer of the argumentation annotation from a corpus in one language to a corpus in another language.

Aker and Zhang [2] created the first annotated Chinese corpus using the existing English corpora and manually matching claims and premises with parallel Chinese texts.

Eger et al. [5] created corpora in German, French, Spanish, and Chinese using human and machine translations of the PersEssays corpus. Eger et al. also compare the annotation projection and direct transfer strategy.

Sliwa et al. [28] created the first annotated corpus for Arabic and the Balkan language group using parallel corpora and annotation transfer of the English version of the corpus based on classifier training.

Eger et al. [6] examined cross-lingual transfer solving two tasks: sentence-level argumentation mining and automatic morphological tagging. They combined two cross-lingual approaches – direct transfer and projection, eliminating the shortcomings of both methods and combining their strengths.

Toledo-Ronen et al. [31] explored the potential of transfer learning using the multilingual BERT model for argumentation mining in non-English languages, based on English datasets and machine translation. They fine-tuned BERT for argumentation mining tasks using training on a corpus that includes both the original English-language texts and those translated into several languages.

In our study, in contrast to [2] and [28], we obtained performance scores of ADUs classification into two classes – "pro" and "opp" based on traditional machine learning techniques (SVM, Bagging and XGBoost), neural network model (BERT) and their ensemble. In contrast to [5] and [6], we explore the effect of expanding the training dataset and the importance of various groups of features. Unlike [31], we work with the BERT version for one language (Russian) and classify corpora in one language (Russian).

### 2.3 Argumentation mining in Russian

There is very little research on argumentation mining for the Russian language, as opposed to English.

Fishcheva and Kotelnikov [7] created the first annotated corpus for the Russian language based on the translation of the ArgMicro corpus. Also, an automated classification of the "pro" and "opp" sentences was carried out.

Kononenko et al. [12] studied argumentation using the comparative analysis of discourse structures. Various types of argument structures were considered. In order to automatically extract argumentative relations, the analysis of the rhetorical and argumentative annotations was carried out. The experiment was carried out on a corpus of 11 popular science articles from Ru-RSTreebank.

Salomatina et al. [26] described a combined approach to partial extraction of the argumentative structure of text, which can be used if there are no sufficient annotated data to effectively apply machine learning techniques for the direct detection of arguments and their relationships.

In this paper, we develop the approach proposed in [7]. We expand the existing Russian-language corpus with argumentation annotation based on machine translation of the persuasive essays corpus. In contrast to [7], to classify ADUs in the new extended version of the corpus, we use a deep neural network (BERT model) along with traditional machine learning techniques, and also explore the effect of expanding the training dataset.

## 3 Joint Argument Annotation Scheme

The Join Argument Annotation Scheme (JAAS) was developed to enable combined processing of ArgMicro and PersEssays corpora. This annotation scheme is based on those used in ArgMicro and PersEssays.

There are three types of objects in the argument annotation schemes: a topic, a node and an edge [30].

1. The topic is a matter dealt with in a text; with or without indicating point of view (stance).
2. The node is a vertex of the argumentation graph. There are three types of nodes: major claims, regular nodes and neutral nodes.

   - The major claim ("mcl") is a node of the argumentation graph which expresses some point of view related to the topic (conclusion). There may be one (ArgMicro) or two (PersEssays) nodes labeled as "mcl". If there are two major claims then both reflect the same stance.
   - Regular nodes are the nodes of the argumentation graph which provide arguments ("pro" or "opp") related to the major claim.
   - Neutral nodes are the sentences which are not members of the argumentation graph in Persuasive Essays Corpus.

3. The edge is a unit which determines a connection between two nodes. There are five types of edges:

   - support ("sup") – a source node supports a target node;
   - additional support ("add") – two or more source nodes support a target node only if they are taken together;
   - example ("exa") – a source node serves as an example of the support of a target node;
   - rebuttal ("reb") – a source node rebuts a target node;
   - undercut ("und") – a source node attacks the connection (edge) between some two nodes.

In the ArgMicro annotation scheme, the graph nodes represent the propositions: the proponent's nodes and the opponent's nodes. The edges connecting the nodes represent different supporting and attacking moves.

In PersEssays annotation scheme, the sentences are classified as major claims, claims, premises and neutral. The PersEssays annotation scheme has unlabeled connections. We convert unlabeled edge types of PersEssays into three types of edges: "sup", "reb" and "exa". Types of edges in ArgMicro are more variable. Thus, types of edges in JAAS are equivalent to ArgMicro edges.

For illustration purposes, Figures 2 and 3 give an example of graph conversion from ArgMicro and PersEssays to JAAS.

## 4 Materials and Methods

### 4.1 Text corpora

Within this study, we used the Argumentative Microtext Corpus (ArgMicro) [24], [27] and the Persuasive Essay Corpus (PersEssays) [29]. Fishcheva and Kotelnikov [7] showed that the best result among Fishcheva I. N., Goloviznina V. S., Kotelnikov E. V.

the Google Translate, Yandex.Translate and Promt systems was demonstrated by Google Translate in the machine translation of the ArgMicro corpus in English into Russian. Therefore, the PersEssays corpus was also translated into Russian using Google Translate[3]. Then the annotations of both corpora were converted to JAAS. A specific issue when converting PersEssays annotation to JAAS was identifying the "example" ("exa") edge types. To address this issue a two-stage procedure was used. At the first stage, an automatic search was carried out for ADUs containing template phrases such as "for example", "for instance", etc. At the second stage, the selected ADUs were manually checked. If the presence of the "example" relation type in the target corpus is not essential, this procedure can be omitted.

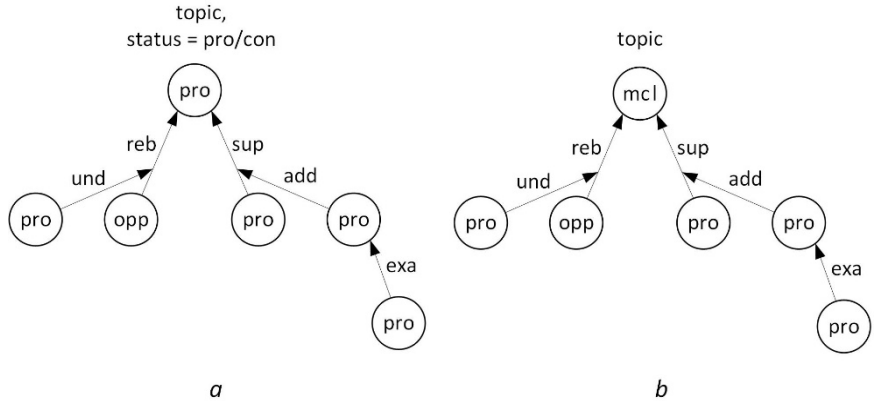

Figure 2: Equivalent graph representation of argumentation structure: a – ArgMicro; b – JAAS

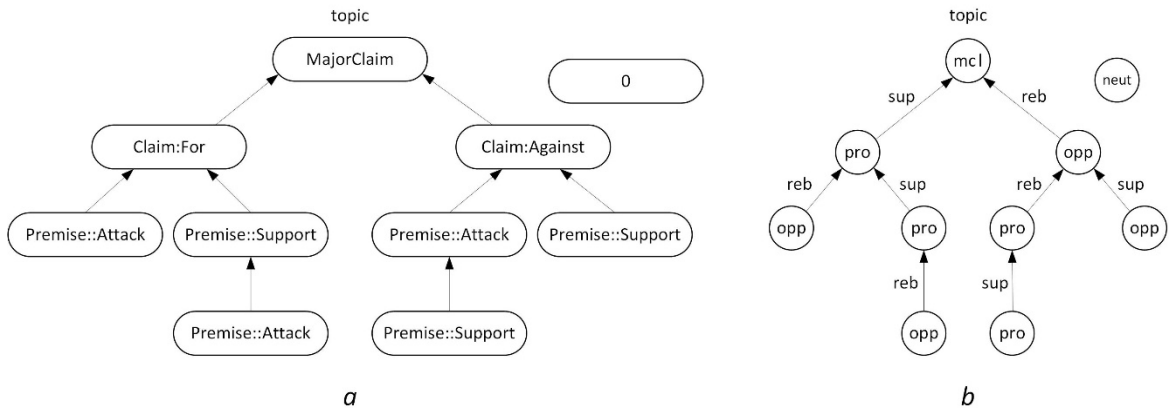

Figure 3: Equivalent graph representation of argumentation structure: a – PersEssays; b – JAAS

Thus, we used the Russian-language versions of the ArgMicro and PersEssays corpora, annotated in accordance with JAAS. The number of texts and ADUs in converted corpora is shown in Table 1. Both individual sentences and parts of sentences can be ADUs in these corpora. For the ArgMicro corpus the inter-annotator agreement by Fleiss kappa is equal to 0.83 (three annotators) [24]. For the PersEssays corpus the inter-annotator agreement by Krippendorff $\alpha_U = 0.72$ for argument components and $\alpha = 0.81$ for argumentative relations [29].

The ArgMicro corpus consists of 1,537 edges of "seg" type, 730 – "sup", 245 – "reb", 140 – "und", 78 – "add", 32 – "exa"; the PersEssays corpus: 7277 – "seg", 5617 – "sup", 715 – "reb", 301 – "exa".

Figure 4 shows an example of text from the ArgMicro corpus in the JAAS, where ADUs, their types ("mcl", "opp", "pro") and relationships between them ("reb", "und", "sup", "add") are indicated.

---

[3] https://translate.google.ru.

| Corpora | Texts | ADUs | | | | |
|---|---|---|---|---|---|---|
| | | pro | opp | mcl | neut | all |
| ArgMicro | 283 | 983 (63.8%) | 253 (16.4%) | 301 (19.5%) | 4 (0.3%) | 1,541 (100%) |
| PersEssays | 399 | 4,599 (63.2%) | 703 (9.7%) | 746 (10.2%) | 1,229 (16.9%) | 7,277 (100%) |
| ArgMicro +PersEssays | 682 | 5,582 (63.3%) | 956 (10.8%) | 1,047 (11.9%) | 1,233 (14.0%) | 8,818 (100%) |

Table 1: Characteristics of text corpora

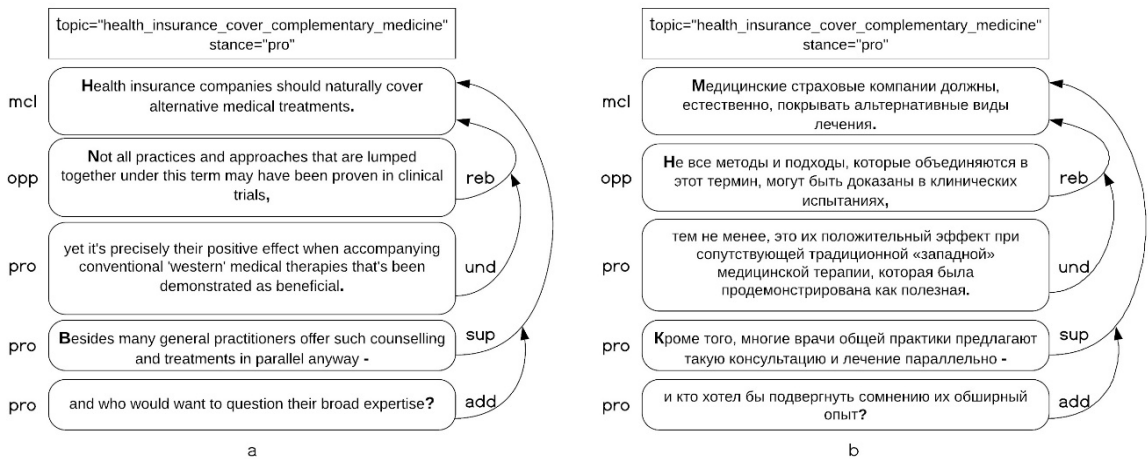

Figure 4: An example of text from the ArgMicro corpus:
*a* – original English-language variant; *b* – Russian-language variant (machine translation)

### 4.2 Features

Fishcheva and Kotelnikov [7] argued that when using traditional machine learning classifiers, the TF.IDF features, the word2vec features and the location of the sentence in the text do not improve the performance of the classifiers. Therefore, in this study, only the following three types of features were considered:
- lexical features – discourse markers ("consequently", "I think", "eventually", etc.) and modal words ("need", "maybe", "necessarily", etc.), including negations, 255 features in total;
- punctuation features – comma, colon, semicolon, question and exclamation marks, 5 features in total;
- morphosyntactic features – N-grams based on parts of speech (nouns, pronouns, verbs, adjectives and adverbs), N = {2, 3, 4}, and grammatical features of verbs: tense, mood, person, 783 features in total.

The preprocessing was carried out on the basis of tokenization and removal of stop words using *nltk*[4], as well as lemmatization using *mystem*[5]. For each ADU, a single vector was formed based on the concatenation of all feature types for the current ADU, as well as all feature types for the previous and next ADUs (if they present) in order to take the context into account.

---
[4] https://www.nltk.org.
[5] https://yandex.ru/dev/mystem.

### 4.3 Traditional machine learning techniques

For training, we used classifiers that gave the best results in [7] – linear Support Vector Machines (SVM), Bagging classifier and Gradient Boosting. The hyperparameters of the latter two classifiers were selected from the following ranges:
- Bagging: number of trees = [50, 100, 200, 500];
- Gradient Boosting: number of trees = [150]; maximum depth of a tree = [2, 8, 20, 30].

We used the SVM and Bagging implementation in *scikit-learn* [22], and for the gradient boosting we used the *XGBoost* library[6].

### 4.4 BERT

Bidirectional Encoder Representations from Transformers (BERT) [4] is a deep neural network language model based on the Transformer architecture [33]. The model is trained on a large text corpus using two tasks: masked words and next sentence prediction. The model is then fine-tuned to solve particular natural language processing tasks. BERT allows bi-directional context-dependent text processing. The model accepts a sequence of tokens (subwords) as input, which is then advanced through several layers of the encoder. The number of layers is 12 ($BERT_{BASE}$) or 24 ($BERT_{LARGE}$). Each layer applies self-attention mechanism and passes the results to the feed-forward network, after which the output of the current layer is fed to the input of the next layer. The encoder output is used as input to a linear classifier with a SoftMax function.

Within this study, the experiments were carried out using the RuBERT model from *DeepPavlov* [[14]]. RuBERT is a multilingual version of $BERT_{BASE}$ (12 layers, hidden size 768, feed-forward hidden size 3,072, and 12 self-attention heads), trained on the Russian-language Wikipedia and the news corpus.

The hyperparameters of the RuBERT model in our experiments were chosen from the following ranges:
- number of epochs = [3, 5, 7];
- learning rate = [$10^{-3}, 10^{-4}, 5 \cdot 10^{-5}, 10^{-5}, 10^{-6}, 10^{-7}$];
- batch size = [4, 8, 16, 32].

## 5 Results and Discussion

### 5.1 Design of experiments

The experiments were carried out in order to get answers to the following questions:
- Q1: What performance of the binary classification of Russian-language ADUs into "pro" and "opp" can be achieved by modern machine learning models?
- Q2: Is it possible to improve the classification performance by expanding the training corpus?
- Q3: What is the significance of different types of features for traditional machine learning classifiers?

To conduct the experiments, we used the ArgMicro and PersEssays corpora, translated into Russian using Google Translate, annotated with JAAS. During the training, only "pro" and "opp" ADUs in these corpora were taken into account; "mcl" and "neut" were ignored (see Table 1).

We studied four variants to create training and test datasets:
- training on the ArgMicro, testing on the ArgMicro;
- training on the PersEssays, testing on the PersEssays;
- training on the ArgMicro and PersEssays, testing on the ArgMicro;
- training on the ArgMicro and PersEssays, testing on the PersEssays.

Due to the small size of the corpora, we used a 5-fold cross-validation for each of the four variants. The partitions were random, stratified by class, and kept the same for all experiments. In addition, in the experiments with RuBERT, with regard to the random initialization of the linear classifier weights, for each of the four variants five runs of the training procedure were carried out.

---

[6] https://xgboost.readthedocs.io.

Because of the strong class imbalance of both corpora, the macro-averaged F1-score was used as the main performance metric. Accuracy and macro-averaged Precision and Recall were also calculated. The results obtained for each experiment were averaged, and the standard deviation for folds was computed.

Nested 3-fold cross-validation was used to fit the hyperparameters. The following hyperparameter values turned out to be optimal for the RuBERT model in most runs:
- training on the ArgMicro: number of epochs – 5, learning rate – $10^{-5}$, batch size – 4;
- training on the PersEssays and training on the joint dataset (ArgMicro and PersEssays): number of epochs – 5, learning rate – $10^{-5}$, batch size – 32.

### 5.2 Results

Table 2 shows the results for the XGBoost classifier, which turned out to be the best in all experiments among other traditional machine learning techniques (SVM and Bagging), as well as the results of RuBERT model. XGBoost results are presented for the full set of features – lexical, punctuation, and morphosyntactic (see Subsection 4.2).

| Train dataset | Test dataset | Model | $F_1$-score | Precision | Recall | Accuracy |
|---|---|---|---|---|---|---|
| ArgMicro | ArgMicro | XGBoost | **0.7921 ±0.0309** | 0.8567 ±0.0437 | **0.7597 ±0.0324** | **0.8819 ±0.0166** |
| | | RuBERT | 0.7441 ±0.0537 | 0.7678 ±0.0414 | 0.7318 ±0.0598 | 0.8468 ±0.0228 |
| ArgMicro+ PersEssays | | XGBoost | 0.7678 ±0.0203 | **0.8583 ±0.0152** | 0.7288 ±0.0204 | 0.8746 ±0.0081 |
| | | RuBERT | 0.7349 ±0.0231 | 0.7691 ±0.0345 | 0.7159 ±0.0237 | 0.8429 ±0.0161 |
| PersEssays | PersEssays | XGBoost | 0.6308 ±0.0191 | **0.7617 ±0.0433** | 0.6009 ±0.0132 | **0.8793 ±0.0073** |
| | | RuBERT | **0.6715 ±0.0339** | 0.7211 ±0.0292 | **0.6469 ±0.0298** | 0.8744 ±0.0088 |
| ArgMicro+ PersEssays | | XGBoost | 0.6510 ±0.0165 | 0.7488 ±0.0303 | 0.6194 ±0.0120 | 0.8791 ±0.0066 |
| | | RuBERT | 0.6665 ±0.0299 | 0.7250 ±0.0230 | 0.6398 ±0.0255 | 0.8752 ±0.0085 |

Table 2: Results of XGBoost and RuBERT:
macro-averaged F1-score, Precision, Recall and Accuracy (Mean ± Std Dev)

### 5.3 Discussion

The best result for the ArgMicro corpus (question Q1) was obtained using XGBoost ($F_1$-score=0.7921) when trained only on the ArgMicro. RuBERT lags far behind ($F_1$-score=0.7441): the ArgMicro corpus includes only 1,236 "pro" and "opp" ADUs, which are not enough for high-quality fine-tuning of the RuBERT, especially considering 5-fold cross-validation. Particularly low is the Precision for RuBERT relative to XGBoost (0.7678 vs. 0.8567).

For the PersEssays corpus, RuBERT produces the best result ($F_1$-score=0.6715). It outperforms XGBoost due to higher Recall (0.6469 vs. 0.6009). The PersEssays corpus (5,302 ADUs) is 4.3 times the size of ArgMicro, and RuBERT is able to train at a level that surpasses traditional machine learning techniques.

Expanding the training dataset (question Q2) by adding PersEssays to ArgMicro in the case of testing on ArgMicro worsens the results for XGBoost (by 0.024) and slightly decreases for RuBERT (by 0.009). Both classifiers lose performance due to macro-averaged Recall, which in turn is reduced due to Recall for the "opp" class: if the PersEssays corpus with a stronger class imbalance is added, it impairs the ability of classifiers to recognize minority class (see Table 3).

When ArgMicro is added to PersEssays, the results are diverse: for XGBoost, the performance improves by 0.02, for RuBERT – almost does not change (decreases by 0.005). When ArgMicro is added, the class imbalance is slightly reduced by increasing Recall for the "opp" class for XGBoost.

XGBoost produces more stable results: the standard deviation of results for folds is lower than for RuBERT (0.0217 vs. 0.0352 on average for all experiments).

The class imbalance in both corpora (the "pro" class is 79.5% in ArgMicro, 86.7% in PersEssays), leads to extremely uneven performance by class. Table 3 shows the performance metrics of the best models in Table 2 by classes.

| Train dataset | Test dataset | Model | Macro $F_1$-score | Class | $F_1$-score | Precision | Recall |
|---|---|---|---|---|---|---|---|
| ArgMicro | ArgMicro | XGBoost | 0.7921 | pro (79.5%) | 0.9286 | 0.8940 | 0.9664 |
|  |  |  |  | opp (20.5%) | 0.6556 | 0.8193 | 0.5529 |
| PersEssays | PersEssays | RuBERT | 0.6715 | pro (86.7%) | 0.9296 | 0.9043 | 0.9565 |
|  |  |  |  | opp (13.3%) | 0.4134 | 0.5379 | 0.3373 |

Table 3: Results of the best classifiers (XGBoost for ArgMicro and RuBERT for PersEssays) for "pro" and "opp" classes: F1-score, Precision and Recall

Table 4 contains the number of ADUs that were classified identically and differently by both classifiers. For example, the column "XGBoost – true, RuBERT – false" shows the number of ADUs that were correctly predicted by XGBoost and incorrectly – by RuBERT.

| Test dataset | Class | XGBoost – true, RuBERT – true | XGBoost – true, RuBERT – false | XGBoost – false, RuBERT – true | XGBoost – false, RuBERT – false | Sum |
|---|---|---|---|---|---|---|
| ArgMicro | all | 975 | 127 | 74 | 60 | 1,236 |
|  | pro | 887 | 66 | 30 | 0 | 983 |
|  | opp | 88 | 61 | 44 | 60 | 253 |
| PersEssays | all | 4,466 | 203 | 187 | 446 | 5,302 |
|  | pro | 4,339 | 139 | 96 | 25 | 4,599 |
|  | opp | 127 | 64 | 91 | 421 | 703 |

Table 4: Results of classification by number of ADUs

The analysis of Table 4 allows us to advance a hypothesis that the ensemble of both models will perform better than the models separately. The rule for predicting the ADU class in the ensemble is as follows: if at least one of the classifiers predicts a minority class "opp", return "opp"; otherwise return "pro". The ensemble results are shown in Table 5 along with the best classifier for the respective corpora.

| Test dataset | Model | $F_1$-score | Precision | Recall | Accuracy |
|---|---|---|---|---|---|
| ArgMicro | XGBoost | 0.7921±0.0309 | **0.8567±0.0437** | 0.7597±0.0324 | **0.8819±0.0166** |
|  | Ensemble | **0.8157±0.0305** | 0.8022±0.0306 | **0.8326±0.0312** | 0.8738±0.0226 |
| PersEssays | RuBERT | 0.6715±0.0339 | **0.7211±0.0292** | 0.6469±0.0298 | **0.8744±0.0088** |
|  | Ensemble | **0.6901±0.0138** | 0.7159±0.0185 | **0.6723±0.0114** | 0.8716±0.0068 |

Table 5: Results of ensemble of classifiers and the best classifiers (XGBoost for ArgMicro and RuBERT for PersEssays): F1-score, Precision and Recall (Mean ± Std Dev)

Table 5 shows that the use of the proposed ensemble allows improving the classification performance by 0.024 for ArgMicro and by 0.019 for PersEssays.

### 5.4 Feature importance

To answer question Q3 about the significance of various types of features, the dependence of the XGBoost classification performance on various combinations of features was investigated:
- lexical features – only lexical features in the previous, the current and the following ADUs;
- all without discourse markers – all features (lexical, punctuation and morphosyntactic) without discourse markers in the previous, the current and the following ADUs;
- all without features of previous ADUs – lexical, punctuation and morphosyntactic features only for the current and the following ADUs;
- all features – a full set of the features in the previous, the current and the following ADUs.

The results are shown in Table 6.

| Train dataset | Test dataset | Lexical features | All without discourse markers | All without features of previous ADUs | All features |
|---|---|---|---|---|---|
| ArgMicro | ArgMicro | 0.8092±0.0273 | 0.5440±0.0098 | **0.8116±0.0355** | 0.7921±0.0309 |
| PersEssays | PersEssays | **0.6534±0.0184** | 0.5140±0.0156 | 0.6284±0.0125 | 0.6308±0.0191 |

Table 6: Results of XGBoost for various combinations of feature types: macro-averaged F1-score (Mean ± Std Dev)

The discourse markers are the most important features, because without these features the performance of the classifier drops drastically. The previous ADU features are the most useless, since the exclusion of these features did not worsen the classifier performance, but allowed obtaining the best result for XGBoost ($F_1$-score=0.8116). Punctuation and morphosyntactic features are not very useful, because when these features are excluded (column "Lexical features"), the result is either close to the best (ArgMicro) or the best (PersEssays).

## 6 Conclusion

Thus, in order to use the ArgMicro and PersEssays corpora combined, the Join Argument Annotation Scheme based on the schemes used in ArgMicro and PersEssays has been proposed. The PersEssays corpus was translated into Russian using Google Translate and made publicly available[7]. We investigated the problem of classifying ADUs into two classes – "pro" and "opp". The experimental study made it possible to answer the questions posed:

Q1: What performance of the binary classification of Russian-language ADUs into "pro" and "opp" can be achieved by modern machine learning models? – The best performance (macro-averaged $F_1$-score) can be achieved by the proposed ensemble of XGBoost and RuBERT: for ArgMicro $F_1$-score=0.8157, for PersEssays $F_1$-score=0.6901. The performance for PersEssays is worse, firstly, due to the fact that the corpus is more imbalanced, and secondly, PersEssays ADUs are longer and more complex than in ArgMicro – the average ADU length in PersEssays is 18.6 tokens vs. 13.8 tokens in ArgMicro.

Q2: Is it possible to improve the performance of classification by expanding the training corpus? – Yes, if the imbalance of the extended corpus is reduced in comparison to the original one. If a less imbalanced ArgMicro corpus was added to a more imbalanced PersEssays, the performance for the XGBoost classifier was slightly higher. In other cases, the performance either did not increase or decreased.

Q3: What is the significance of different types of features for traditional classifiers? – Discourse markers turned out to be the most important features; features of previous ADUs have minimal impact on the performance of the classifier.

The urgent tasks to be solved in further research are, firstly, expanding the range of Russian-language corpora with argumentative annotation both based on the translation of existing corpora in other languages, and using annotation by people; secondly, the study of the performance of the argumentation mining for new corpora.

## Acknowledgements

The reported study was jointly financed by the German Academic Exchange Service (DAAD) and the Ministry of Science and Higher Education of the Russian Federation within the "Michail Lomonosov" programme (2021).

---

[7] https://github.com/kotelnikov-ev/PersEssays_Russian.